\documentclass{article}
\PassOptionsToPackage{numbers,compress}{natbib}
 \usepackage[preprint]{neurips_2020}
\usepackage[utf8]{inputenc}
\usepackage[T1]{fontenc}
\usepackage[pagebackref=true,breaklinks=true,letterpaper=true,colorlinks,bookmarks=false]{hyperref}
\usepackage{amsmath}
\usepackage{amssymb}
\usepackage{bbm}
\usepackage{booktabs}
\usepackage{graphbox}
\usepackage{graphicx}
\usepackage{multirow}
\usepackage{subcaption}
\usepackage{xcolor}

\makeatletter
  \newcommand\figcaption{\def\@captype{figure}\caption}
  \newcommand\tabcaption{\def\@captype{table}\caption}
\makeatother

\newcommand{\tabincell}[2]{\begin{tabular}{@{}#1@{}}#2\end{tabular}}
\newcommand{\srcim}{I^{\mathrm{s}}}
\newcommand{\targetim}{I^{\mathrm{t}}}
\newcommand{\predim}{\hat{I}^{\mathrm{t}}}
\newcommand{\depth}{D}
\newcommand{\mask}{M}

\newcommand{\etal}{et al.}

\graphicspath{{./}{../}{../../}}

\begin{document}
\title{Moving SLAM\@: Fully Unsupervised Deep Learning in Non-Rigid Scenes}

\author{
Dan Xu\\
HKUST\\
{\tt\small danxu@cse.ust.hk}
\and
\textbf{Andrea Vedaldi}\\
University of Oxford\\
{\tt\small vedaldi@robots.ox.ac.uk}
\and
\textbf{João F. Henriques} \\
University of Oxford\\
{\tt\small joao@robots.ox.ac.uk}
}

\maketitle
\begin{abstract}
We propose a method to train deep networks to decompose videos into 3D geometry (camera and depth), moving objects, and their motions, with no supervision.
We build on the idea of view synthesis, which uses classical camera geometry to re-render a source image from a different point-of-view, specified by a predicted relative pose and depth map.
By minimizing the error between the synthetic image and the corresponding real image in a video, the deep network that predicts pose and depth can be trained completely unsupervised.
However, the view synthesis equations rely on a strong assumption: that objects do not move.
This rigid-world assumption limits the predictive power, and rules out learning about objects automatically.
We propose a simple solution: minimize the error on small regions of the image instead.
While the scene as a whole may be non-rigid, it is always possible to find small regions that are approximately rigid, such as inside a moving object.
Our network can then predict different poses for each region, in a sliding window from a learned dense pose map. This represents a significantly richer model, including 6D object motions, with little additional complexity.
We achieve very competitive performance on unsupervised odometry and depth prediction on KITTI\@.
We also demonstrate new capabilities on EPIC-Kitchens, a challenging dataset of indoor videos, where there is no ground truth information for depth, odometry, object segmentation or motion --- yet all are recovered automatically by our method.
\end{abstract}

\raggedbottom

\section{Introduction}\label{s:intro}

It is a long-standing goal of computer vision to achieve a holistic understanding of a visual scene: that is, to decompose it into meaningful elements that together explain the full visual input~\cite{hartley2003multiple}.
This goal is also at the heart of representation learning, which is concerned with extracting representations from data that generalize well for multiple tasks~\cite{zamir2018taskonomy,xu2018PAD-Net}.
For example, a representation that was trained for object detection will invariably ignore details that are crucial for other tasks that are not object-centric, such as monocular depth estimation.
As such, it is desirable to learn models that are not narrowly-scoped to a single task~\cite{caruana1997multitask}, though it is not always clear how to do so without an increased annotation burden for each additional task.

A recent line of work that promises to achieve both goals at once --- a holistic scene understanding without demanding additional annotations --- is unsupervised learning by view synthesis~\cite{garg2016unsupervised,zhou2017unsupervised}.
It cleverly combines two predictions (i.e. tasks), relative camera pose estimation and depth estimation, to re-render an image from another point-of-view.
By synthesizing a source frame from a video into another target frame, we obtain the supervision necessary (by comparing the synthetic and the real image) to enable end-to-end learning.

It is natural to ask if more tasks can be integrated into this framework, and intuitively the answer is yes.
Each additional task that contributes to the image synthesis, by capturing an additional element of the visual world, should increase the fidelity of the model and achieve a lower reconstruction error.
In this work, we propose to add the unsupervised tasks of object segmentation, and object 6D motion estimation.
Since our primary interest is still the recovery of 3D geometry and motion, however, we do not seek, as done by some prior work~\cite{casser2019depth, tosi2020distilled, guizilini2020semantically}, ``semantic'' objects.
Instead, we decompose the image into regions that are likely to be characterized by a well defined rigid motion, and learn those automatically, by optimizing for the same view synthesis objective.
Our technical contribution is a locally-rigid model that supports crisp object boundaries through segmentation, and its efficient implementation by reusing a commonly-available but underused ``tiling'' operator.
This results in an efficient lightweight model that transforms the original rigid-world model (where changes between frames are characterized only by a relative camera pose and depths)~\cite{zhou2017unsupervised} to a non-rigid model (where moving objects are also taken into account).

Experiments show that our proposed model is much more expressive, by making predictions that are outside the capabilities of the original model (namely segment moving objects and calculate their 6D motions).
We show qualitative results of our neural networks successfully learning to segment non-rigid objects (hands and household objects), and recover accurate depth maps, in EPIC-Kitchens~\cite{Damen2020Collection}, a large-scale indoors dataset that has not been used in this context before due to its challenging nature.
We also demonstrate that the unsupervised segmentation cues and non-rigid model are beneficial for the previously-considered tasks.
Our model achieves a new state-of-the-art result on KITTI~\cite{Geiger2013IJRR} in monocular depth estimation (88.9\% accuracy at $\delta < 1.25$), as well as visual odometry (0.011/0.010 ATE), from purely unsupervised data (see sec.~\ref{s:results} for details).

The paper is organized as follows: sec.~\ref{s:related} discusses relevant literature; sec.~\ref{s:background} introduces the view synthesis framework; sec.~\ref{s:method} presents our method, and sec.~\ref{s:results} the results; sec.~\ref{s:conclusion} concludes our paper.

\section{Related work}\label{s:related}

Self-supervised learning~\cite{caruana1997promoting,self-supervised-survey2019} has gathered significant attention recently, since it promises to achieve unsupervised learning by reusing standard elements from supervised learning (e.g. architectures, loss functions), in relatively intuitive configurations.
It can generally be described as the task of predicting one part of the input data given only another part.
Successful examples include predicting the spatial relationship between two image regions~\cite{doersch2015unsupervised}, colorizing images~\cite{zhang2016colorful}, predicting audio from video~\cite{nagrani2018seeing,owens2016ambient}, and predicting inertial measurement unit (IMU) data (odometry) from video~\cite{agrawal2015learning}.
A related approach is to use synthetic transformations, such as predicting the rotation of an image~\cite{gidaris2018unsupervised} or whether a video is played in reverse~\cite{wei2018learning}.
Given the large body of literature on self-supervised learning, we refer interested readers to a recent survey~\cite{self-supervised-survey2019}.
Within self-supervised learning, video prediction is a relatively popular task. It consists of generating a subset of a video given the remaining video.
Several works have focused on video generation with neural networks~\cite{srivastava2015unsupervised}, by proposing multiple-hypothesis loss functions~\cite{rupprecht2017learning}, causal convolutions such as the PixelCNN for video~\cite{kalchbrenner2017video}, and generative models such as adversarial networks or variational auto-encoders~\cite{videogen2019}.
While successful, these approaches often produce blurry or physically-implausible predictions, and the image generation models are not interpretable.

Approaching this problem by view synthesis~\cite{zhou2017unsupervised} is relatively recent, but draws from traditional works in visual geometry and Simultaneous Location And Mapping (SLAM)~\cite{scaramuzza2011visual,davison2007monoslam,Sheng2019Unsupervised}.
Instead of using a neural network for generation, it uses a differentiable warp (image deformation) to transform a source frame into a target frame, minimizing the reconstruction error. The task of the neural network is then to predict physically-interpretable quantities, such as 3D geometry, depth and poses, that are used to compute the image warp. This physical model of the world is useful in itself, and can be used for downstream tasks (such as visual odometry or 3D reconstruction).
This line of work was pioneered by Garg et al.~\cite{garg2016unsupervised}, who considered stereo pairs or pairs of frames with known pose. This was further developed to include predicted poses by Zhou et al.~\cite{zhou2017unsupervised} (SfMLearner), and concurrently by Vijayanarasimhan et al.~\cite{sfmnet2017} (SfM-Net), who also considered multiple layers to account for objects. The SfMLearner emphasized no supervision (as opposed to mixed modes of supervision like SfM-Net), and a simple architecture, so it forms the basis of our work. We describe it in detail in sec.~\ref{s:background}. Further developments include improving the pose estimate by a direct visual odometry method (essentially second-order gradient descent)~\cite{wang2018learning}, and adding a FlowNet that can refine the image warp by fine-grained optical flow estimates~\cite{yin2018geonet}. Other proposed improvements are stereo inputs~\cite{godard2017unsupervised,li2018undeepvo} and probabilistic outputs~\cite{klodt2018supervising}, bundle adjustment to warp stored key frames instead of recent frames~\cite{yang2018deep}, adversarial training~\cite{pilzer2018unsupervised}, and feature computation in 3D space~\cite{guizilini2019packnet}.

\section{Unsupervised learning by view synthesis}\label{s:background}

We begin by describing the canonical self-supervised setup for learning monocular depth estimation, inspired by early methods such as SfMLearner~\cite{zhou2017unsupervised} (sec.~\ref{s:related}).
Assume that we are given a pair of images $(\srcim,\,\targetim)$, respectively the \emph{source} and \emph{target}, usually extracted as nearby frames from a video.
If the scene is Lambertian and unchanged between frames (rigid-world assumption) and if we discount occlusions, the target image $\predim$ can be predicted from knowledge of the source image $\srcim$, the depth map $\depth$, and the relative camera pose $P$ between the two views:
\begin{equation}\label{e:synth}
\predim=\Psi\left(\srcim,\,P,\,\depth\right)
\end{equation}
Synthesizing the target image amounts to using projective geometry~\cite{hartley2003multiple} to find which pixels correspond in the two views and then transport their intensities from source to target.
Namely, the synthesis function $\Psi$ is a warp that linearly interpolates each pixel of $\srcim$ at coordinates $(u,v)$ to $(u',v')$ according to the projective equations:
\begin{equation}\label{e:warp}
    \left[u',v',1\right]^{T}=KPZ\left(\depth_{uv}\right)K^{-1}\left[u,v,1\right]^{T},
\end{equation}
where $K$ is the camera intrinsics matrix, $\depth_{uv}$ retrieves the depth at pixels $(u,v)$, and $Z(\depth_{uv})$ is the transformation matrix that translates any input 3D point along the $z$ axis by this depth.
We can extract a supervisory signal for depth and pose by comparing the measured and predicted target image:
\begin{equation}\label{e:sfm-objective}
    L_\textrm{rigid} =\frac{1}{|\mathcal{D}|}\sum_{(\srcim,\,\targetim)\in\mathcal{D}}\left|\targetim-\Psi\left(\srcim,\,P,\,\depth\right)\right|_{\mask}
\end{equation}
where $\mathcal{D}$ is a training dataset of image pairs, $M$ is a mask that tells which pixels can be explained by the model and $|v|_\mask=\sum_k \mask_k |v_k|$, i.e.~the $L^1$ norm weighted by a mask $\mask$.

Of course, the pose $P$, depth $\depth$ and mask $M$ must be obtained somehow.
This is the role of the pose, depth and mask networks:
\begin{equation}\label{e:nets}
    P=\rho\left(\targetim,\srcim|\omega\right),\quad
    d=\phi\left(\targetim|\omega\right),\quad
    \mask=\epsilon\left(\targetim,\srcim|\omega\right)
\end{equation}
with parameters $\omega$.
The last layer of the mask network is a sigmoid, to constrain outputs to the $[0, 1]$ range.
Back-propagating the error in eq.~\ref{e:sfm-objective} to the networks' parameters $\omega$ is what allows end-to-end learning.
Since predicting $\mask=0$ is a trivial minimum of eq.~\ref{e:sfm-objective}, a regularization term is added, penalizing the distance between $\mask$ and a target of 1:\footnote{The regularization loss used in~\cite{zhou2017unsupervised} was the cross-entropy, which is not a material difference.}
\begin{equation}
L_{\textrm{reg}}=
\left\Vert M - 1\right\Vert ^{2}
\end{equation}
The overall objective is then $\min_\omega \, L_{\textrm{rigid}} + \lambda L_{\textrm{reg}}$, with a regularization weight $\lambda$.

\paragraph{Implementation.}

Although this is the main objective, Zhou et al.~\cite{zhou2017unsupervised} mention a few details that improve performance:
(1) a smoothness loss (TV-norm), which penalizes the L1 norm of the second-order gradients of the depth maps $\depth$;
(2) repeating the objective for multiple scales of the input image;
(3) using several source images $\srcim$ to predict their poses jointly, instead of one at a time.


\section{Proposed method}\label{s:method}

As we mentioned in sec.~\ref{s:intro}, the main limitation of the method from sec.~\ref{s:background} is that it assumes a rigid world.
Our main goal then is to augment it to also account for freely-moving, potentially non-rigid objects.
Note that since most pixels in common scenes correspond to a static background, it is undesirable to have a fully non-rigid model, as that would afford too many degrees-of-freedom and thus be prone to overfitting (a hypothesis that we verify in sec.~\ref{s:results}).
For this reason, we segment the pixels into 2 categories: background (with a global rigid model), and objects (with a local, non-rigid model).
Together, they fully explain all pixels of the image.
The required segmentation network is trained unsupervised as part of the overall objective, and so object segmentation is obtained ``for free''.
We will now describe these three elements: non-rigid and rigid models, and segmentation.

\begin{figure}
    \centering
    \includegraphics[width=\textwidth]{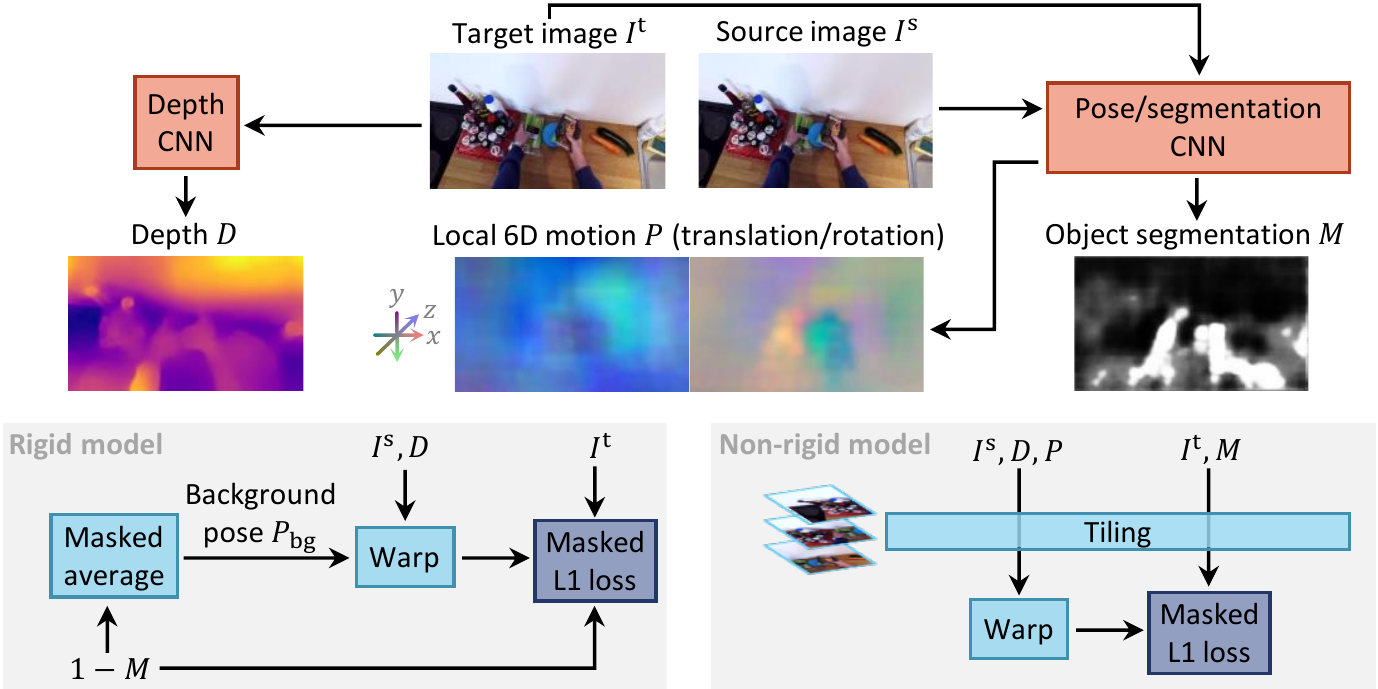}
    \caption{Overview. \textbf{Top panel:} From a pair of video frames, CNNs predict 3 maps: depth, local 6D motion, and foreground segmentation. Higher depths are darker; the axes of motion translation and rotation (XYZ) are encoded as RGB channels; and foreground segmentation is white.
    \textbf{Bottom-left:} We model background pixels as a rigid object, with global 6D motion obtained by averaging their predicted motions.
    A L1 reconstruction error w.r.t. target image $\targetim$ allows learning.
    \textbf{Bottom-right:} Foreground pixels are similar, but only \emph{locally-rigid}; inputs are passed through a \emph{tiling} operator, dividing them into small patches. This allows motions per object/patch, unlike the fully rigid model.
    }\label{f:proposed}
\end{figure}

\subsection{Locally-rigid scene model}\label{s:local-poses}

The starting point for our method is the simple fact that, although a scene is not globally rigid, it usually is \emph{locally} rigid.
It is always possible to make the rigid-world assumption essentially correct, by narrowing down the view to a rigid region (e.g. background, a rigid object, or a smaller region within a non-rigid object).
This seems to suggest that the previous method (sec.~\ref{s:background}) can express a non-rigid model, by focusing on smaller regions of the image at a time instead of full images.
We can then predict a different relative pose prediction for each region, instead of a single pose $P$.
We thus propose to average the objective (eq.~\ref{e:sfm-objective}) over a set of regions $\mathcal{R}$, extracted with a sliding window:
\begin{equation}\label{e:locally-rigid}
    L_{\textrm{non-rigid}}=
    \frac{1}{|\mathcal{D}||\mathcal{R}|}\sum_{(\srcim,\,\targetim)\in\mathcal{D}}\sum_{r\in\mathcal{R}}\left|\targetim-\Psi\left(\srcim,\,P_{r},\,\depth\right)\right|_{\mask\,\cdot\,\mathbf{1}_{r}},
\end{equation}
where we element-wise multiply ($\cdot$) the mask $M$ with an indicator function $\mathbf{1}_{r}$, which is 1 inside the region $r$ and 0 outside of it.
$P_r$ denotes the corresponding pose, extracted from a pose map $P$ at the center of the region.
The dense map of pose predictions $P$ fits naturally into the CNN-based architecture, in the same way as the depth predictions $D$.

While it achieves our goal of a locally-rigid model, eq.~\ref{e:locally-rigid} is very inefficient (by summing over many zeros), and an efficient implementation would require customizing the differentiable warp operator $\Psi$.
However, we can ensure an easy implementation with no modification to $\Psi$ by using an operator that extracts patches in a sliding window, and concatenates them as samples before they are passed on to $\Psi$.
This \emph{tiling} operator, also known as {\tt im2row}, is used in some implementations of convolutions.\footnote{One example of its use for convolutions is in the Caffe deep learning library. In PyTorch, for example, it is implemented as the {\tt unfold} operation.}
We can write it succinctly, for an input with $c$ channels and spatial dimensions $(w,\,h)$, as $t:\,\mathbb{R}^{c\times w\times h}\mapsto\mathbb{R}^{b\times c\times k\times k}$:
\begin{equation}\label{e:tiling}
    t_{i+jw,\,:,\,:,\,:}(X)=X_{:,\,si:si+k,\,sj:sj+k}
\end{equation}
where $:$ is the tensor slice operator, the patches are $k \times k$, and the stride is $s$.
Eq.~\ref{e:locally-rigid} then becomes:
\begin{equation}\label{e:non-rigid}
L_{\textrm{non-rigid}}=L_{\textrm{rigid}}\left(\mathcal{D}',\,\mathrm{vec}(P),\,t(\depth),\,t(\mask)\right),
\quad\mathcal{D}'=\left\{ \left(t(\srcim),\,t(\targetim)\right):\,(\srcim,\,\targetim)\in\mathcal{D}\right\},
\end{equation}
which effectively amounts to applying the tiling operator $t$ to all image-sized inputs of the original rigid-model objective (eq.~\ref{e:sfm-objective}), and vectorizing the pose map $P\in\mathbb{R}^{6 \times w \times h}$ so that the spatial dimensions correspond to the batch dimension ($\mathrm{vec}(P)\in\mathbb{R}^{wh \times 6}$).
Since the kernel size $k$ is unknown, in practice we repeat the objective for several values of $k$, corresponding to different object sizes.
Our proposed inclusion of the tiling operators is illustrated in fig.~\ref{f:proposed}.


\subsection{Object segmentation}\label{s:segmentation}

As discussed in sec.~\ref{s:method}, a fully non-rigid model contains too many degrees-of-freedom.
Another issue is that regions defined by square sliding windows (eq.~\ref{e:locally-rigid}) are too coarse to accurately delineate objects' boundaries.
We propose to solve both issues at once by using the predicted mask $M$ to partition the pixels into moving objects ($M \simeq 1$) and static background ($M \simeq 0$):
\begin{equation}\label{e:segmentation}
L_{\textrm{segm}}=
L_{\textrm{non-rigid}}+L_{\textrm{rigid}}(\mathcal{D},P_{\textrm{bg}},\depth,1-\mask),\quad
P_{\textrm{bg}}=\tfrac{1}{wh}\sum\nolimits_{ij}^{wh}(1-\mask_{ij})P_{ij},
\end{equation}
where $P_{\textrm{bg}}$ is the background pose, obtained by averaging the pose of all background pixels.
Unfortunately, since the non-rigid model is more expressive than the rigid one, assigning all pixels to the former will always attain a lower error.
To prevent this trivial solution, we modify the regularization term (eq.~\ref{e:sfm-objective}) to encourage a constant area of foreground pixels in a training batch.
We consider that the top 10\% of the predictions $M$ correspond to pixels with moving objects ($M \simeq 1$), and the rest to background ($M \simeq 0$), penalizing the distance to these target values:
\begin{equation}\label{e:foreground-reg}
L_{\textrm{reg'}}=
\left\Vert \mathrm{sort}(\mathrm{vec}(M))-u\right\Vert ^{2}
\end{equation}
where sort operates in descending order, and $u$ is a vector where the first 10\% of the elements are 1 and the rest are 0.
This ``constant area'' soft constraint is inspired by a similar approach used for visualizing salient regions in deep network interpretability~\cite{fong2019understanding}, and we found it to be an effective strategy to ensure a correct proportion of object and background pixels.
Our overall objective is then $\min_\omega \, L_{\textrm{segm}} + \lambda L_{\textrm{reg'}}$, in analogy to sec.~\ref{s:background}.
An illustration of the objective is shown in fig.~\ref{f:proposed}.

\section{Experiments}\label{s:results}

We conducted a series of experiments to validate the effectiveness of our approach.
The tasks we focused on were visual odometry, monocular depth estimation, and 6D motion segmentation.

\begin{figure*}[!t]
    \centering
    \includegraphics[width=.99\textwidth]{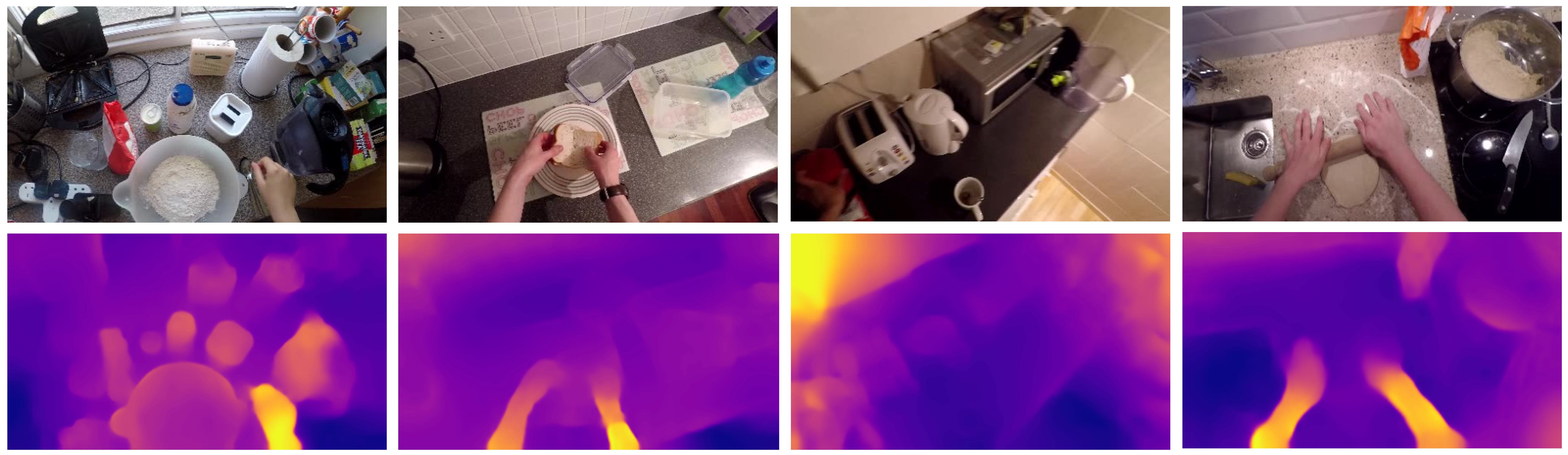}
    \caption{Qualitative examples of our unsupervised depth estimation on the EPIC-Kitchens dataset. Despite the fast, non-rigid motions, we can recover detailed structures, such as the tabletop objects.}
    \label{fig:epic_visual}
\end{figure*}

\paragraph{Datasets.}
We evaluate the proposed approach on two different large-scale datasets. The first one is the challenging autonomous driving benchmark \textbf{KITTI}~\cite{Geiger2013IJRR}.
For the depth estimation, we use the training split defined by Eigen et al.~\cite{eigen2014depth}.
For evaluating the performance of the visual odometry, we use the KITTI Odometry dataset, training on sequences $00-08$ and testing on $09-10$.
For the second dataset, we used \textbf{EPIC-Kitchens}~\cite{Damen2020Collection}, which is collected under various indoor kitchen scenarios, and is the largest dataset for egocentric vision. It contains 32 kitchens crossing 4 cities, totalling 55 hours of video. It captures rich non-rigid dynamic motions. Some examples are shown in Fig.~\ref{fig:epic_visual}. As the original frame rate of the videos is 60 FPS, to reduce the redundancy, we sample the dataset at every 4 frames, resulting in a dataset of around 120k images. Among them, 100k images are used for training, and the rest for testing. The dataset does not provide ground-truth depth, camera poses and intrinsics. We learn to recover all of them by unsupervised end-to-end learning.

\paragraph{Training setup.}
Our training procedure is exactly the same as for the SfMLearner~\cite{zhou2017unsupervised}, except with the non-rigid model we propose (sec.~\ref{s:method}).
We used a ResNet-50 as the backbone CNN, and apply the same TV-norm (sec.~\ref{s:background}) to both depth ($\depth$) and pose ($P$) maps.
The only other improvements are depth mean normalizaton and backbone initialization according to Gordon et al.~\cite{gordon2019depth}, as well as disabling the multi-scale prediction, which does not seem beneficial.
For brevity, we do not describe the full setup here, but a self-contained description can be found in the supp.~material~(Appendix~A).

\begin{figure*}[!t]
    \centering
    \includegraphics[width=.99\textwidth]{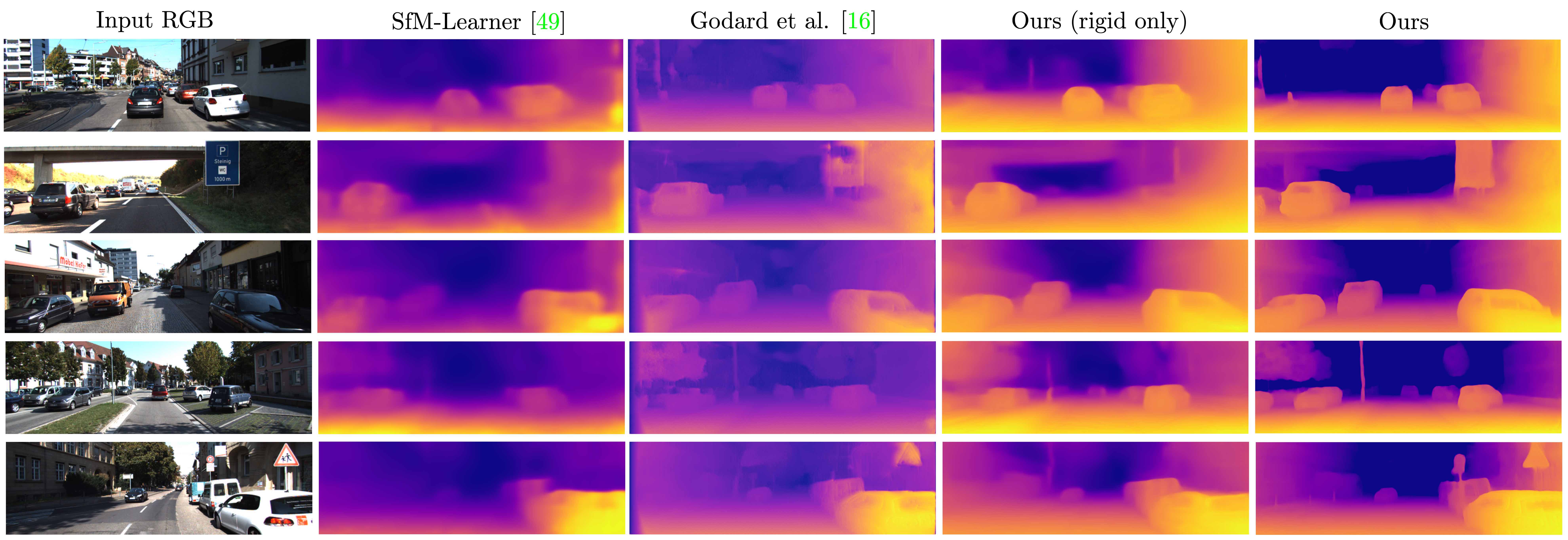}
    \vspace{-3pt}
    \caption{Visualisation of depth estimates for our method, including an ablation without the non-rigid component, and two other methods, including one with access to stereo information~\cite{godard2017unsupervised}.
    Our method seems more accurate, capturing moving cars and fine details such as traffic signs and poles.
    }
    \label{fig:kitti_visual}
    \vspace{-2pt}
\end{figure*}

\begin{figure*}[htb]
\vspace{0pt}
\begin{minipage}{0.395\textwidth}
  \footnotesize
    \centering
    \setlength\tabcolsep{4pt}
\resizebox{0.99\linewidth}{!} {
    \begin{tabular}{l|c|c}
\toprule[1.5pt]
\multirow{2}{*}{Method} & \multicolumn{2}{c}{\tabincell{c}{Absolute Trajectory Error}} \\ \cline{2-3}
& sequence 09 & sequence 10 \\
\midrule
Mean Odometry & 0.032 $\pm$ 0.026 & 0.028 $\pm$ 0.023 \\
ORB-SLAM (short) & 0.064 $\pm$ 0.141 & 0.064 $\pm$ 0.130 \\
ORB-SLAM (full) & 0.014 $\pm$ 0.008 & 0.012 $\pm$ 0.011 \\
Zhou et al.~\cite{zhou2017unsupervised} & 0.021 $\pm$ 0.017 & 0.020 $\pm$ 0.015 \\
DF-Net~\cite{zou2018df} & 0.017 $\pm$ 0.007 & 0.015 $\pm$ 0.009 \\
Monodepth2~\cite{gordon2019depth} & 0.017 $\pm$ 0.008 &  0.015 $\pm$ 0.010\\
Zhou et al.~\cite{zhou2017unsupervised} (new) & 0.016 $\pm$ 0.009 & 0.013 $\pm$ 0.009 \\
Bian et al.~\cite{bian2019unsupervised} & 0.016 $\pm$ 0.007 & 0.015 $\pm$ 0.015 \\
Klodt~\etal~\cite{klodt2018supervising} & 0.014 $\pm$ 0.007 & 0.013 $\pm$ 0.009 \\
Mahjourian et al.~\cite{mahjourian2018unsupervised} & 0.013 $\pm$ 0.010 & 0.012 $\pm$ 0.011 \\
EPC++~\cite{luo2018every} & 0.013 $\pm$ 0.007 & 0.012 $\pm$ 0.008 \\
GeoNet~\cite{yin2018geonet} & 0.012 $\pm$ 0.007 & 0.012 $\pm$ 0.009 \\
CC~\cite{ranjan2019competitive} & 0.012 $\pm$ 0.007 & 0.012 $\pm$ 0.009 \\
\midrule
Ours & \textbf{0.011 $\pm$ 0.005} & \textbf{0.010 $\pm$ 0.007} \\
\bottomrule[1.5pt]
\end{tabular}
}
\tabcaption{Absolute Trajectory Error (ATE) on the KITTI odometry test split averaged over all $5$-frame snippets.
Our approach outperforms all others, including a traditional SLAM pipeline.
}
\label{tab:ate_pose}
\end{minipage}
\hfill
\begin{minipage}{0.585\textwidth}
\vspace{-4.5pt}
    \centering
    \begin{subfigure}[t]{0.61\linewidth}
        \centering
        \includegraphics[height=1.97in]{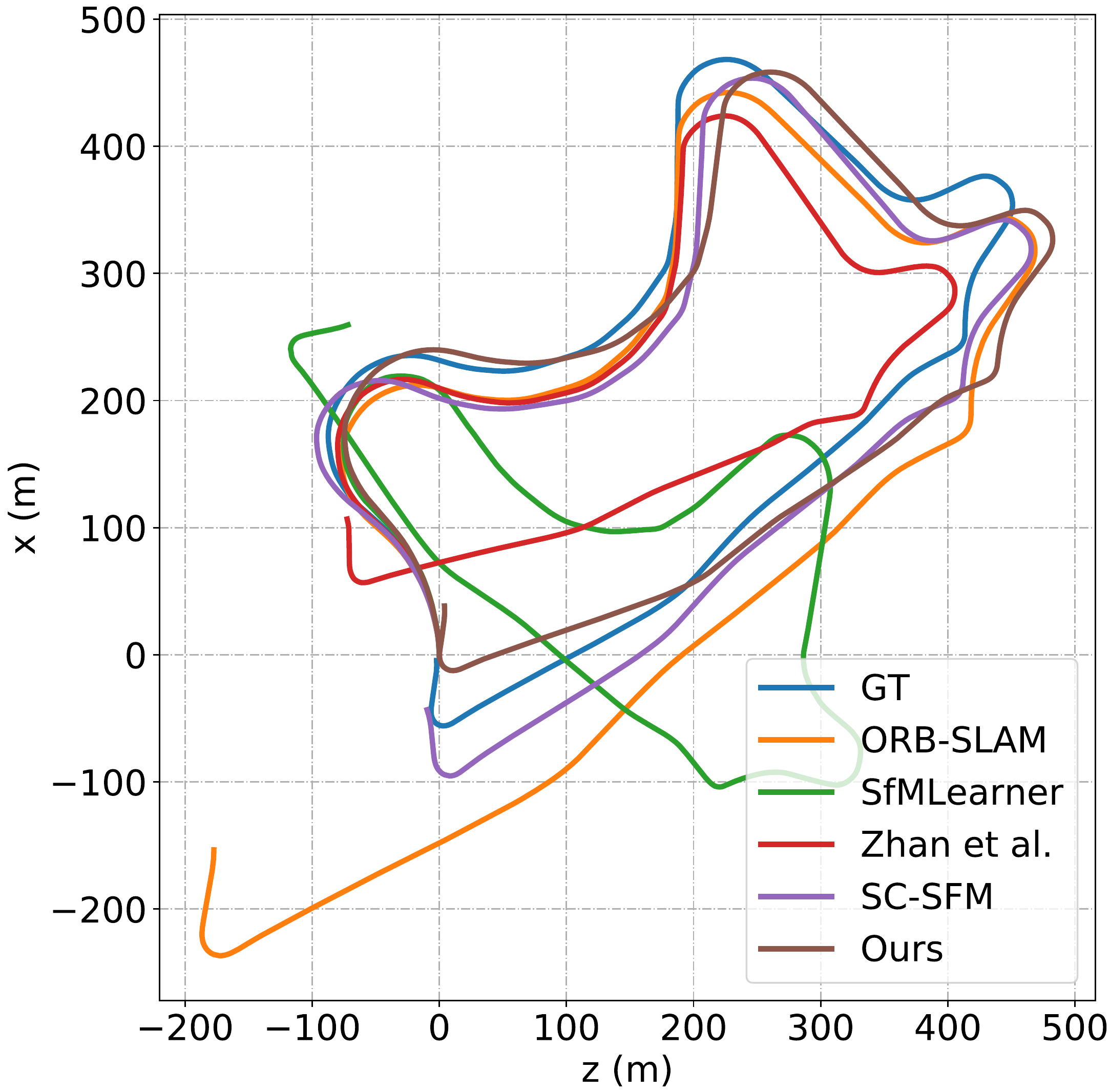}
        \caption{Testing sequence 09}
    \end{subfigure}%
    \begin{subfigure}[t]{0.39\linewidth}
        \centering
        \includegraphics[height=1.97in]{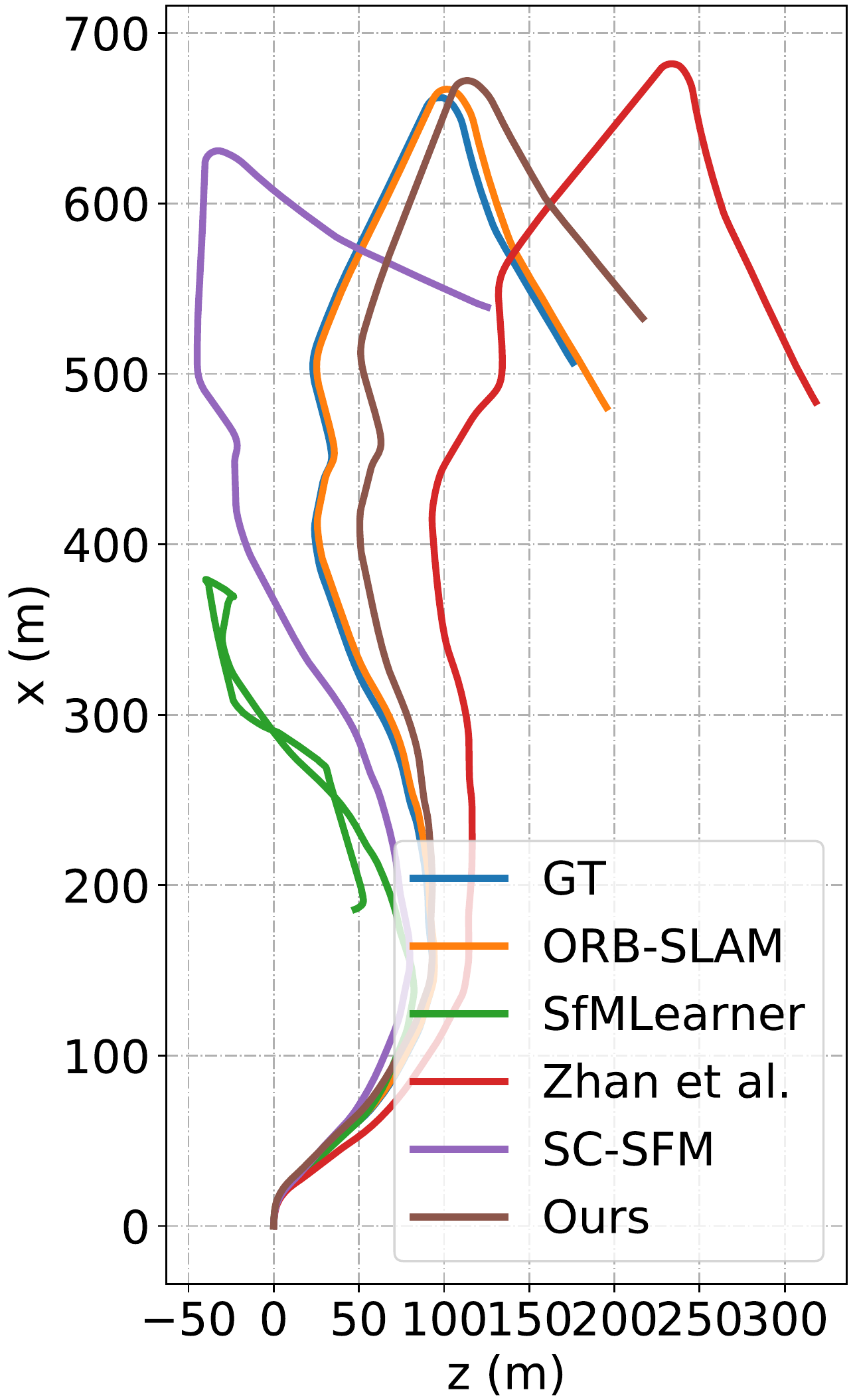}
       \caption{Testing sequence 10}
    \end{subfigure}
    \vspace{-3pt}
    \figcaption{Qualitative state-of-the-art comparison of the visual odometry results on the full testing sequences 09 and 10 of the KITTI odometry dataset.}
\label{fig:kitti_odometry_09_10}
  \end{minipage}%
  \vspace{-8pt}
\end{figure*}

\paragraph{Pose estimation performance.}
Our sequential pose estimation (visual odometry) performance is reported in table~\ref{tab:ate_pose}.
We compare our method with a traditional monocular SLAM system, ORB-SLAM (full)~\cite{mur2015orb}, as well as its local version, ORB-SLAM (short), which uses $5$-frame snippets.
We also compare with the SfMLearner~\cite{zhou2017unsupervised} and several recent proposals~\cite{zou2018df,mahjourian2018unsupervised,yin2018geonet,bian2019unsupervised,klodt2018supervising,ranjan2019competitive}.
As can be seen in table~\ref{tab:ate_pose}, our method outperforms all the other methods.
This includes a traditional SLAM pipeline that draws from many years of careful engineering and manual tuning (ORB-SLAM).
It is worth mentioning that our method is not explicitly trained for visual odometry, yet it is a useful by-product of training.
Regarding the deep learning based approaches, our camera motion estimator outperforms the original SfMLearner~\cite{zhou2017unsupervised} by a large margin, and recent competing approaches by narrower but still significant margins. We include the standard deviations over several runs in table~\ref{tab:ate_pose} for additional context.

\paragraph{Qualitative results on visual odometry.}
We visualize the predicted camera trajectories in fig.~\ref{fig:kitti_odometry_09_10}a and \ref{fig:kitti_odometry_09_10}b, for several algorithms, on two KITTI test sequences (09 and 10, respectively).
All trajectories are registered w.r.t. the ground truth as standard~\cite{zhan2018unsupervised}.
It is apparent that the trajectory predicted by our method very accurately follows the ground truth.
While other methods can also get close to the ground truth trajectory (SC-SFM in fig.~\ref{fig:kitti_odometry_09_10}a and ORB-SLAM in \ref{fig:kitti_odometry_09_10}b), no other can achieve the same performance simultaneously on both settings.


\paragraph{Depth estimation performance.}
We report the result for depth estimation in table~\ref{overall_KITTI}. The columns are relative error and its square, root-mean-squared error (RMSE) and its logarithm, as well as the accuracy at 3 given depth thresholds.
We include several state-of-the-art approaches in the comparison, including supervised methods~\cite{eigen2014depth,liu2015deep} and unsupervised stereo-based methods~\cite{godard2017unsupervised,garg2016unsupervised}, which are not comparable but present an informative upper bound on performance.
We do not include in the comparison works with significantly different protocols, such as 3DPackNet~\cite{guizilini2019packnet} which uses higher-resolution images and 3D CNNs, and other works with higher amounts of supervision. These interesting developments are complementary to ours, and thus outside of the scope of this paper.
All methods are compared with raw ground truth depth, following recent protocols~\cite{gordon2019depth, guizilini2019packnet}.
We can observe that our method achieves the best performance out of all unsupervised methods, even beating several supervised ones, on almost all metrics.
We show two variants of our method, using ResNet-18 and ResNet-50 backbone CNNs, which highlights that the performance gains are not solely due to an increase in capacity, since most methods have comparable backbones.
On the other hand, it also reveals that our model is expressive enough to afford some gains when increasing the capacity to ResNet-50. More importantly, our method achieves clearly better results compared with two recent works, i.e. Struct2Depth~\cite{casser2019depth} and Tosi et al.~\cite{tosi2020distilled}, which use explicit semantic labels to guide the learning of object motion, demonstrating the benefits of unsupervised segmentation.

\begin{table*}[!t]
\centering
\caption{
Quantitative comparison of depth estimation performance among several methods from the literature, on the KITTI raw dataset (Eigen~\etal~\cite{eigen2014depth} testing split).
We show some supervised (denoted `{D}') and stereo (`\textcolor{blue}{S}') methods for reference, but a fair comparison is only with monocular methods (`\textcolor{red}{M}') trained with video (\textcolor{cyan}{`V'}).
Our method outperforms others in most metrics, even when using a ResNet-18 backbone with much fewer parameters.
}
\vspace{3pt}
\setlength\tabcolsep{8pt}
\resizebox{0.97\linewidth}{!} {
\begin{tabular}{l|c|cccc|ccc}
\toprule[1.5pt]
\multirow{2}{*}{Method} & \multirow{2}{*}{Setting} & \multicolumn{4}{c|}{\tabincell{c}{Error (lower is better)}} & \multicolumn{3}{c}{\tabincell{c}{Accuracy (higher is better)}} \\\cline{3-9}
 &  & rel & sq rel & rmse & rmse (log) & $\delta < 1.25$ & $\delta < 1.25^2$ & $\delta < 1.25^3$ \\\midrule
Eigen et al.~\cite{eigen2014depth}    & \textcolor{red}{M} + \textcolor{black}{D} & 0.203 & 1.548 & 6.307 & 0.282 & 0.702 & 0.890 & 0.958 \\
Liu et al.~\cite{liu2015deep}    &   \textcolor{red}{M} + \textcolor{black}{D} & 0.202 & 1.614 & 6.523 & 0.275 & 0.678 & 0.895 & 0.965 \\

AdaDepth~\cite{adadepth} &  \textcolor{blue}{S} & 0.203 & 1.734 & 6.251 & 0.284 & 0.687 & 0.899 & 0.958 \\
Garg et al.~\cite{garg2016unsupervised} &  \textcolor{blue}{S} & 0.169 & 1.080  & 5.104   & 0.273  &  0.740 & 0.904  &  0.962   \\
Zhan et al.~\cite{zhan2018unsupervised} &  \textcolor{blue}{S} & 0.144 & 1.391 & 5.869 & 0.241  & 0.803 & 0.933 & 0.971 \\
MS-CRF~\cite{xu2018monocular} & \textcolor{red}{M} + \textcolor{black}{D}  & 0.125 & 0.899 & 4.685 & - & 0.816 & 0.951 & 0.983 \\
Godard et al.~\etal~\cite{godard2017unsupervised} & \textcolor{blue}{S} & {0.124} & {1.076} & {5.311} & {0.219} & {0.847} & {0.942} & {0.973} \\
Kuznietsov et al.~\cite{kuznietsov2017semi} &  \textcolor{blue}{S} + \textcolor{black}{D} & {0.113} & {0.741} & {4.621} & {0.189} & {0.862} & {0.960} & {0.986} \\

\midrule \midrule
Zhou et al.~\cite{zhou2017unsupervised}  &  \textcolor{red}{M} + \textcolor{cyan}{V}
& {0.208}   &  {1.768}  & {6.858}  &  {0.283} & {0.678}   & {0.885}   & {0.957}  \\
Yang et al.~\cite{yang2018unsupervised} & \textcolor{red}{M} + \textcolor{cyan}{V} &  0.182 & 1.481 & 6.501 & 0.267 & 0.725 & 0.906 & 0.963 \\
Mahjourian~\cite{mahjourian2018unsupervised} & \textcolor{red}{M} + \textcolor{cyan}{V} & 0.163 & 1.240 & 6.220 &  0.250 & 0.762 & 0.916 & 0.968 \\
Geonet (ResNet)~\cite{yin2018geonet} & \textcolor{red}{M} + \textcolor{cyan}{V} & 0.155 & 1.296 & 5.857 & 0.233 & 0.793 & 0.931 & 0.973 \\
Wang et al.~\cite{wang2017learning} & \textcolor{red}{M} + \textcolor{cyan}{V} & 0.151 &
 1.257 & 5.583 & 0.228  & 0.810 & 0.936 & 0.974 \\
DF-Net~\cite{zou2018df} & \textcolor{red}{M} + \textcolor{cyan}{V} & 0.150 & 1.124 & 5.507 & 0.223 & 0.806 & 0.933 & 0.973 \\
Struct2Depth~\cite{casser2019depth} & \textcolor{red}{M} + \textcolor{cyan}{V} & 0.141 & 1.026 & 5.291 & 0.215 & 0.816 & 0.945 & 0.979 \\
CC (ResNet)~\cite{ranjan2019competitive} &  \textcolor{red}{M} + \textcolor{cyan}{V} & 0.140 & 1.070 & 5.326 & 0.217 & 0.826 & 0.941 & 0.975 \\
Bian et al.~\cite{bian2019unsupervised} & \textcolor{red}{M} + \textcolor{cyan}{V} & 0.128 & 1.047 & 5.234 & 0.208 & 0.846 & 0.947 & 0.976 \\
Gordon et al.~\cite{gordon2019depth} & \textcolor{red}{M} + \textcolor{cyan}{V} & 0.128 & 0.959 & 5.230 & -- & -- & -- & -- \\

Tosi et al.~\cite{tosi2020distilled} & \textcolor{red}{M} + \textcolor{cyan}{V} & 0.125 & 0.805 & 4.795 & 0.195 & 0.849 & 0.955 & 0.983 \\

Monodepth2~\cite{gordon2019depth} & \textcolor{red}{M} + \textcolor{cyan}{V} & 0.115 & 0.882 & \textbf{4.701} & 0.190 & 0.879 & 0.961 & 0.982 \\

\midrule
\midrule
Ours (ResNet-18) & \textcolor{red}{M} + \textcolor{cyan}{V} & 0.105 & 0.889 & 4.780 & 0.182 & 0.884 & 0.961 & 0.982 \\

Ours (ResNet-50) & \textcolor{red}{M} + \textcolor{cyan}{V}
&\textbf{0.103} & \textbf{0.881} & {4.763} & \textbf{0.179} & \textbf{0.889} & \textbf{0.964} & \textbf{0.984} \\
\bottomrule[1.5pt]
\end{tabular}
}
\label{overall_KITTI}
\vspace{-5pt}
\end{table*}



\paragraph{Qualitative results on depth.}
Fig.~\ref{fig:kitti_visual} shows a direct comparison of the depth estimates from our method, the most comparable baseline by Zhou et al.~\cite{zhou2017unsupervised}, and a stereo-based method~\cite{garg2016unsupervised}.
Our method recovers much finer details, compared to a rigid-world model (4th column). It is also interesting to compare the level of detail with the stereo method (3rd column), which achieves better error metrics (table~\ref{overall_KITTI}), but has relatively inconsistent fine details. Ours seems to strike a good balance between capturing the high-level layout and small-scale features.
We also show qualitative results of our method in EPIC-Kitchens, for which there is no ground truth, in fig.~\ref{fig:epic_visual}.
Despite the challenges of the quick camera and object motions in this setting, the quality of the recovered depths is apparent, making it possible to identify individual small objects and fine geometry.





\begin{figure*}[!t]
    \centering
    \includegraphics[width=0.33\textwidth]{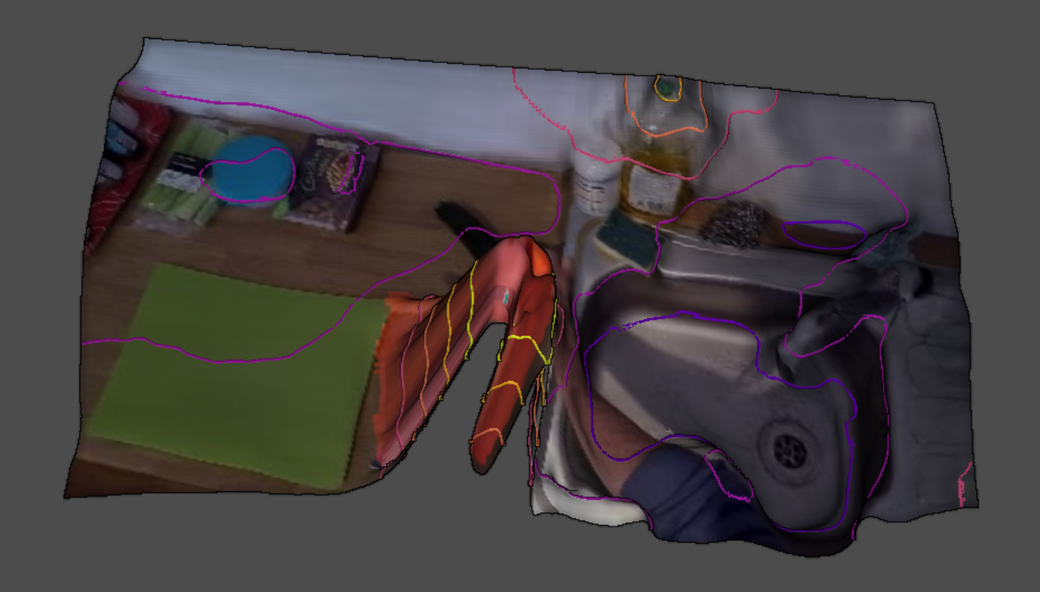}%
    \includegraphics[width=0.33\textwidth]{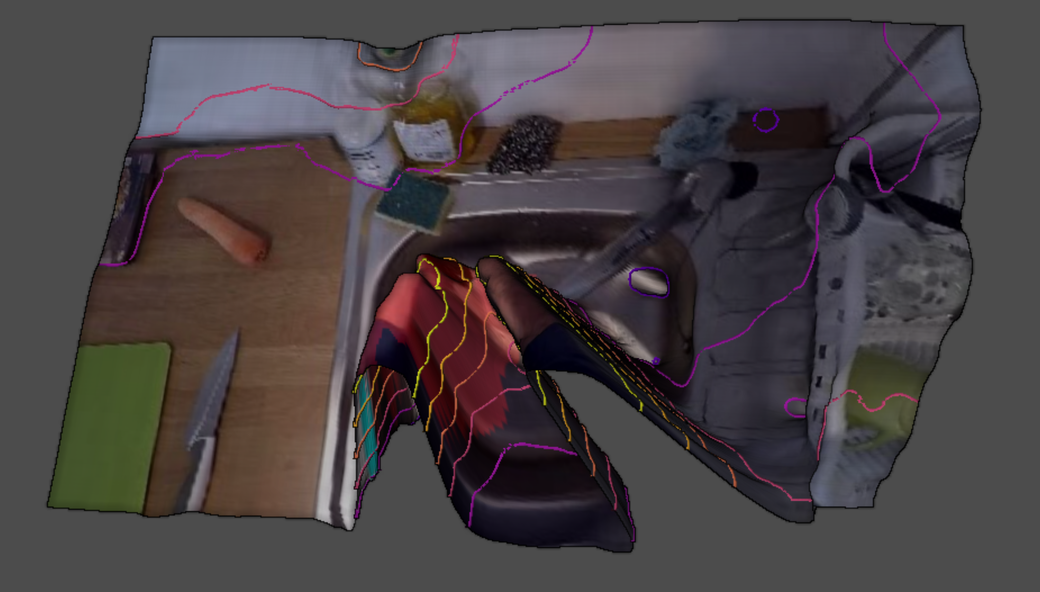}%
    \includegraphics[width=0.33\textwidth]{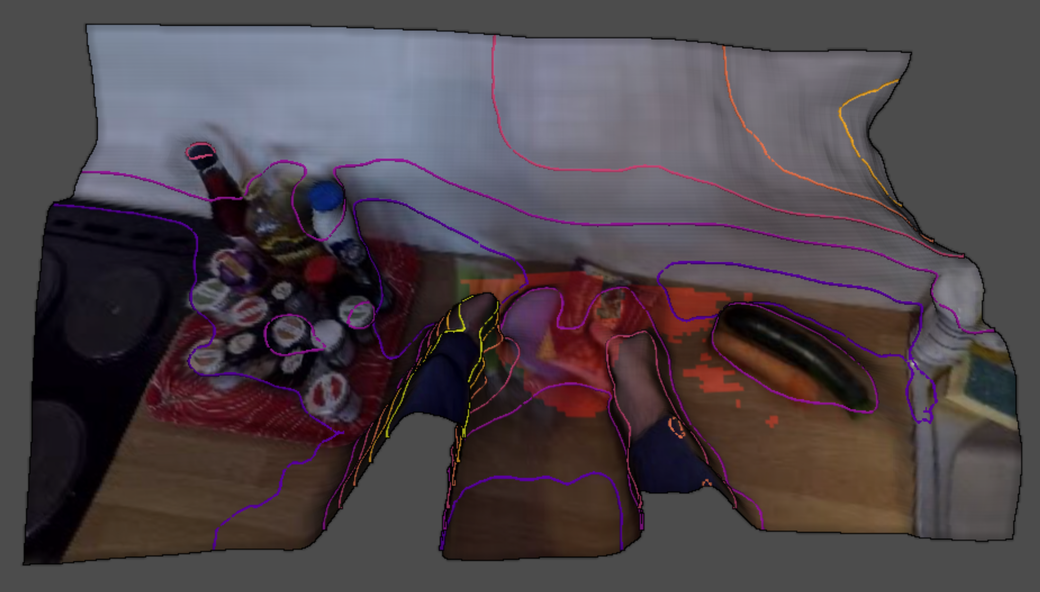}
    
\caption{Predicted 3D mesh and segmentations on the EPIC-Kitchens test set.
Contours of constant height are shown as lines (higher values are brighter). Note the correct geometry of the sink (left panel) and bottle cap (right panel). A failure mode is also visible on the right panel (back wall is distorted in the top-right).
Object masks are tinted red; normally corresponding to moving hands (also higher/nearer to the camera) or manipulated objects.
}
\label{fig:motion}
\vspace{-3pt}
\end{figure*}

\begin{figure}[!t]
\begin{minipage}{0.44\textwidth}
\centering
\footnotesize
\renewcommand\arraystretch{1.265}
\resizebox{1.0\linewidth}{!} {
\begin{tabular}{l|ccc}
\toprule[1.5pt]
Method
& rel & sq rel & rmse \\\midrule
SfMLearner~\cite{zhou2017unsupervised} & 0.208   &  1.768  & 6.858   \\
SfMLearner + our non-rigid model  & 0.173 & 1.302 & 6.252\\
+ regularization (TV-norm)  & \textbf{0.165}  & \textbf{1.301} &  \textbf{6.225}\\\midrule\midrule
Our rigid model (eq.~\ref{e:sfm-objective}) & 0.121 & 1.115 & 5.285 \\
Our non-rigid model (eq.~\ref{e:non-rigid}) & 0.108 & 0.899 & 4.788 \\
+ regularization (TV-norm) & 0.107 & 0.895 & 4.786\\
+ segmentation (eq.~\ref{e:segmentation}-\ref{e:foreground-reg}) (full system) & \textbf{0.105} & \textbf{0.889} & \textbf{4.780}\\
\bottomrule[1.5pt]
\end{tabular}
}
\tabcaption{Ablation study on monocular depth estimation (error).
See sec.~\ref{s:results} for details.
}\label{tab:ablation_KITTI}
\end{minipage}
\hfill
\begin{minipage}{0.54\textwidth}
    \centering
\centering
\renewcommand\arraystretch{1.2}
\resizebox{0.98\linewidth}{!}%
{
\begin{tabular}{l|c|c}
\toprule[1.5pt]
Method & seq. 09 (ATE) & seq. 10 (ATE) \\
\midrule
SfMLearner\cite{zhou2017unsupervised} & 0.021 $\pm$ 0.017 & 0.020 $\pm$ 0.015 \\
SfMLearner + our non-rigid model & 0.012 $\pm$ 0.007 & 0.011 $\pm$ 0.008\\
+ regularization (TV-norm) & \textbf{0.011 $\pm$ 0.005} & \textbf{0.011 $\pm$ 0.007} \\\midrule\midrule
Our rigid model (eq.~\ref{e:sfm-objective}) & {0.016 $\pm$ 0.012} & {0.015 $\pm$ 0.013}  \\
Our non-rigid model (eq.~\ref{e:non-rigid}) & {0.013 $\pm$ 0.006} &  {0.011 $\pm$ 0.010} \\
+ regularization (TV-norm) & {0.012 $\pm$ 0.006} &  
{0.011 $\pm$ 0.009} \\
+ segmentation (eq.~\ref{e:segmentation}-\ref{e:foreground-reg}) (full system) & \textbf{0.011 $\pm$ 0.005} & \textbf{0.010 $\pm$ 0.007} \\
\bottomrule[1.5pt]
\end{tabular}
}
\tabcaption{Ablation study on visual odometry (Absolute Trajectory Error and std. dev.). See sec.~\ref{s:results} for details.
}\label{tab:ablation_pose}
  \end{minipage}
  \vspace{-10pt}
\end{figure}

\paragraph{Qualitative results on object discovery and motion prediction.}
Since our proposed method can predict relative poses (6D motion) densely for the whole image and separate foreground pixels (with independent motions) from static background pixels (with a single coherent motion), it should be able to discover moving objects in the image automatically.
We validate this idea by binarizing the mask $\mask$ (with a threshold of 0.7), and overlaying it on the original data.
The resulting segmentations in EPIC-Kitchens can be visualized in fig.~\ref{fig:motion}, along with the projected 3D geometry.
We can observe the correct segmentation and clustering of the hands, and of objects as they are being held.
We show more visualizations of the motion prediction in the suppl. material (Appendix B).

\paragraph{Ablative study.}
Although our model adds relatively few elements to view synthesis, it is important to know their relative impact on performance.
We show the performance of our system with the addition of each element in tables~\ref{tab:ablation_KITTI}-\ref{tab:ablation_pose}.
It is apparent that the biggest boost comes from adding our non-rigid model, especially in visual odometry performance.
We also show a similar breakdown, but \emph{adding} our improvements to the original SfMLearner system~\cite{zhou2017unsupervised}, which has a weaker backbone CNN and no depth normalization (see Appendix A for details).
This shows that the benefits of our non-rigid model are complementary to other improvements, yet very significant.

\section{Conclusion}\label{s:conclusion}

We have proposed a simple approach to view synthesis for self-supervised learning, which leverages a successful rigid-world model and augments it with a locally-rigid model instead.
This is done by strategic placement of tiling operators within the network, which is both efficient and produces highly consistent depth and pose estimates.
Our change enables unsupervised object discovery by motion segmentation, and allows associating 3D pose and 6D motion to different objects in the scene.
We also demonstrated superior performance in unsupervised visual odometry and monocular depth estimation over comparable baselines.
We hope that our contribution enables a new class of non-rigid SLAM algorithms, that continue the trend towards truly holistic scene understanding.

\newpage

\section{Broader impact}

Our contributions improve the automatic understanding of 3D geometry, including camera and object motions through space. The most widespread applications of this technology nowadays are self-driving vehicles and robots meant to navigate human spaces (including indoors), which have a positive impact for society.

Given that the kind of perception we study has little semantic content, focusing instead on geometry, there is limited potential for harmful biases. It is conceivable that machines intended to cause harm on purpose could be built using computer vision technology, and that 3D environment perception would be one small part of such devices. This is a problem that we strongly believe that society as a whole needs to work through with appropriate regulations, and that in the case of 3D perception, the benefits (e.g. self-driving vehicle technology) outweigh the avoidable risks.

As for unintentional failure cases, we must point out that monocular depth estimation systems are inherently dependent on past experience to make accurate predictions in a given environment. As such, it is important that safety-critical systems (e.g. collision avoidance) do not depend solely on monocular estimates of depth since they may fail in unfamiliar situations; redundant systems (e.g. stereo, other sensors) are necessary.

{\small
\bibliographystyle{ieee_fullname}
\bibliography{egbib}
}

\end{document}